\pdfoutput=1
\documentclass[letterpaper]{article} 
\usepackage{aaai23}  
\usepackage{times}  
\usepackage{helvet}  
\usepackage{courier}  
\usepackage[hyphens]{url}  
\usepackage{graphicx} 
\usepackage{natbib}  
\usepackage{caption} 
\usepackage{multirow}
\usepackage{booktabs}
\usepackage{adjustbox}
\usepackage{amsfonts}
\usepackage{amsmath}
\usepackage{algorithm}
\usepackage{algorithmic}
\usepackage[switch]{lineno}
\usepackage{color}

\usepackage{newfloat}
\usepackage{listings}
\DeclareCaptionStyle{ruled}{labelfont=normalfont,labelsep=colon,strut=off} 
\floatstyle{ruled}
\newfloat{listing}{tb}{lst}{}
\floatname{listing}{Listing}
\title{Cross-Domain Graph Anomaly Detection via \\Anomaly-aware Contrastive Alignment}

\author {
    Qizhou Wang,\textsuperscript{\rm 1}
    Guansong Pang\footnote{Corresponding author: Guansong Pang (gspang@smu.edu.sg).}, \textsuperscript{\rm 2}
    Mahsa Salehi, \textsuperscript{\rm 1}
    Wray Buntine, \textsuperscript{\rm 3,1}
    Christopher Leckie, \textsuperscript{\rm 4}
}
\affiliations {
    \textsuperscript{\rm 1} Monash University\\
    \textsuperscript{\rm 2} Singapore Management University\\
    \textsuperscript{\rm 3} VinUniversity\\
    \textsuperscript{\rm 4} The University of Melbourne\\
    \{qizhou.wang, mahsa.salehi, wray.buntine\}@monash.edu, gspang@smu.edu.sg, caleckie@unimelb.edu.au
}
\usepackage{bibentry}

\begin{document}

\maketitle

\begin{abstract}
Cross-domain graph anomaly detection (CD-GAD) describes the problem of detecting anomalous nodes in an unlabelled target graph using auxiliary, related source graphs with labelled anomalous and normal nodes. Although it presents a promising approach to address the notoriously high false positive issue in anomaly detection, little work has been done in this line of research.
There are numerous domain adaptation methods in the literature, but it is difficult to adapt them for GAD
due to the unknown distributions of the anomalies and the complex node relations embedded in graph data.
To this end, we introduce a novel domain adaptation approach, namely Anomaly-aware Contrastive alignmenT (ACT), for GAD. ACT is designed to jointly optimise: (i) \textit{unsupervised contrastive learning} of normal representations of nodes in the target graph, and (ii) \textit{anomaly-aware one-class alignment} that aligns these contrastive node representations and the representations of labelled normal nodes in the source graph, while enforcing significant deviation of the representations of the normal nodes from the labelled anomalous nodes in the source graph.
In doing so, ACT effectively transfers anomaly-informed knowledge from the source graph to learn the complex node relations of the normal class for GAD on the target graph without any specification of the anomaly distributions.
Extensive experiments on eight CD-GAD settings demonstrate that our approach ACT achieves substantially improved
detection performance over 10 state-of-the-art GAD methods. Code is available at https://github.com/QZ-WANG/ACT.
\end{abstract}

\section{Introduction}
Detection of nodes that deviate significantly from the majority of nodes in a graph is a key task in graph anomaly detection (GAD).
It has drawn wide research attention due to its numerous applications in a range of domains such as intrusion detection in cybersecurity, fraud detection in fintech and malicious user account detection in social network analysis. 
There are many shallow and deep methods \cite{akoglu2015graph,pang2021deep} that are specifically designed, or can be adapted for GAD. However, they are fully unsupervised approaches and often have notoriously high false positives due to the lack of knowledge about the anomalies of interest.

\begin{figure}[t]
    \centering
    \includegraphics[width=0.85\columnwidth]{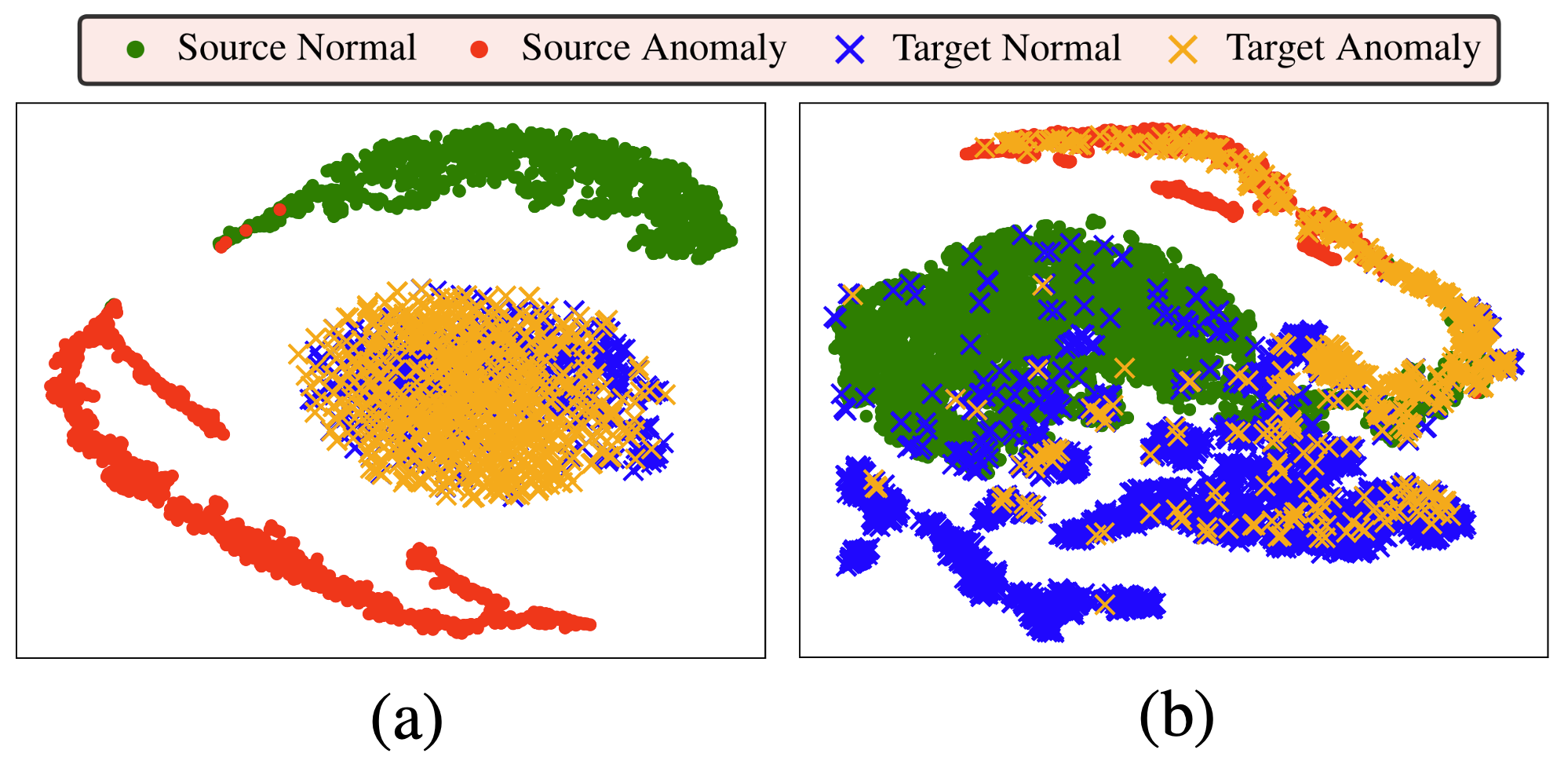}
    \caption{t-SNE visualisation of a CD-GAD dataset 
    before (a) and after (b) our anomaly-aware contrastive alignment.
    Compared to (a) where the two domains show clear discrepancies in different aspects like anomaly distribution, in (b) our domain alignment approach effectively aligns the normal class, while pushing away the anomalous nodes in both source and target domains from the normal class. 
    }
    \label{fig:intuition}
\end{figure}
We instead explore cross-domain (CD) anomaly detection approaches to address this long-standing issue. CD-GAD describes the problem of detecting anomalous nodes in an unlabelled target graph using auxiliary, related source graphs with labelled anomalous and normal nodes. The ground truth information in the source graph can provide important knowledge of true anomalies for GAD on the target graph when such supervision information from the source domain can be properly adapted to the target domain. The detection models can then be trained in an anomaly-informed fashion on the target graph, resulting in GAD models with substantially improved anomaly-discriminative capability, and thus greatly reducing the detection errors.
Although such CD approaches can be a promising solution, little work has been done in this line of research.

There are numerous unsupervised domain adaptation (UDA) methods in the literature \cite{uda_sur_diane}, but it is difficult to adapt them for GAD due to some unique challenges in GAD. 
The first challenge is that the distribution of different anomalies can vary within a dataset and across different datasets, and thus, the anomaly distribution often remains unknown in a target dataset. This challenges the popular assumption in UDA that the source and target domains have similar conditional probability distributions.
Secondly, graph data contains complex node relations due to its topological structure and node attribute semantics, leading to substantial discrepancies in graph structures (e.g., node degree distribution and graph density) and attribute spaces (e.g., feature dimensionality size) across different graph datasets (see Figure \ref{fig:intuition}(a) for an example). These significant domain gaps in the raw input render many UDA methods ineffective, since they require more homogeneous raw input to effectively adapt the domain knowledge (e.g., pre-trained feature representation models can be directly applied to both source and target domains to extract relevant initial latent representations).

To address these two challenges, we introduce a novel domain adaptation approach, namely Anomaly-aware Contrastive alignmenT (ACT), for GAD. ACT is designed to jointly optimise: (i) \textit{unsupervised contrastive learning} of normal representations of nodes in the target graph, and (ii) \textit{anomaly-aware one-class alignment} that aligns these contrastive node representations and the representations of labelled normal nodes in the source graph data, while enforcing significant deviation of the representations of the normal nodes from the labelled anomalous nodes in the source graph. In doing so, ACT effectively transfers anomaly-informed knowledge from the source graph to enable the learning of the complex node relations of the normal class for GAD on the target graph without any specification of the anomaly distributions, as illustrated in Figure \ref{fig:intuition}(b). We also show that after our domain alignment, self-labelling-based deviation learning can be leveraged on the domain-adapted representations of the target graph to refine the detection models for better detection performance.

In summary, our main contributions are as follows:
\begin{itemize}
    \item We propose a novel approach, named anomaly-aware contrastive alignment (ACT), for CD-GAD. It synthesises anomaly-aware one-class alignment and unsupervised contrastive graph learning to learn anomaly-informed detection models on target graph data, substantially reducing the notoriously high false positives due to the lack of knowledge about true anomalies.
    \item We propose the use of self-labelling-based deviation learning on the target graph after the domain alignment to further refine our detection model, resulting in significantly enhanced detection performance.
    \item Large-scale empirical evaluation of ACT and 10 state-of-the-art (SOTA) competing methods is performed on eight real-world CD-GAD datasets to justify the superiority of ACT. These results also establish important performance benchmarks in this under-explored area.
\end{itemize}

\section{Related Work}
\subsection{Graph Anomaly Detection}
GAD methods typically adopt unsupervised learning due to the scarcity of labelled anomalies \cite{jia_wu_gad_survey}. Earlier non-deep-learning-based methods employ various measures \cite{coda, amen, anomalous, ryder} to identify anomalies. Recent GAD methods predominantly use Graph Neural Networks (GNNs) due to their strong learning capacity and are shown to be more effective. \citet{domi} and \citet{ggan} employed graph auto-encoders to define anomaly scores using reconstruction error. Contrastive learning \cite{cola}, adversarial learning \cite{ggan}, and other representation learning approaches \cite{meng_jiang_error, adone, endash} have been explored for GAD.
However, they are unsupervised methods and focused on single-domain GAD.
Limited work has been done on CD-GAD. Two most related studies are \cite{meta-gdn,cmd}. \citet{meta-gdn} adapts a meta-learning approach to address the problem, while \citet{cmd} combine a graph autoencoder and adversarial learning for CD-GAD.
However, they suffer from limitations such as parameter sharing of cross-domain feature learners and unstable performance in the domain alignment.

\subsection{Unsupervised Domain Adaptation}
UDA aims to leverage labelled source data to improve similar tasks in an unlabelled domain.
A popular approach is to reduce domain discrepancies, measured by some predefined metrics such as MMD \cite{mmd, long2016unsupervised} and Wasserstein Distance \cite{WDGRL,swd}. Adversarial learning is also widely used by UDA methods \cite{ganin2015, tzeng2015sim, adda, bousmalis2017unsupervised, hoffman2018cycada, saito2017adversarial,xiao2021dynamic}, which learns domain-invariant representations in a competing training scheme. 
Some recent methods focus on class-wise alignment \cite{xie2018centriod,saito2}. 
These approaches have been recently adapted to graph data, e.g., by adversarial graph learning \cite{dane,pan_www, wu2022attraction} or graph proximity preserved representation learning \cite{guda_adversarial}.
Nevertheless, these methods are primarily designed for CD settings with class-balanced data and relatively small domain discrepancy, rendering them inapplicable for GAD.

\section{ACT: The Proposed Approach}
\begin{figure}
    \centering
    \includegraphics[width=\columnwidth]{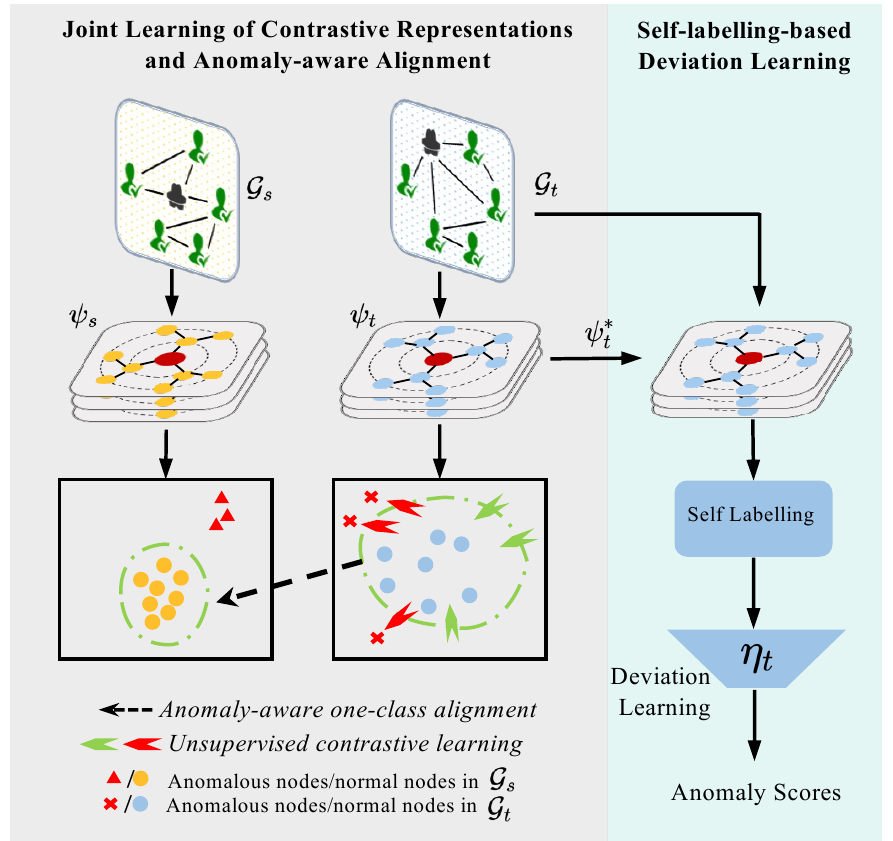}
    \caption{Overview of our proposed approach ACT.}
    \label{fig:fw}
\end{figure}

\subsection{Problem Statement}
We consider unsupervised CD-GAD on attributed graphs. Let $\mathcal{G} = (\mathcal{V}, \mathcal{E}, \mathbf{X})$ be an attributed graph with $n$ nodes, where $\mathcal{V}$, $\mathcal{E}$, and $\mathbf{X} \in \mathbb{R}^{n \times d}$ are its node set, edge set and feature matrix, respectively. 
In the unsupervised CD setting, in addition to a target graph $\mathcal{G}_t = (\mathcal{V}_t, \mathcal{E}_t, \mathbf{X}_t \in \mathbb{R}^{n_t \times d_t})$ with $n_s$ nodes without any class labels, a labelled source graph $\mathcal{G}_s = (\mathcal{V}_s, \mathcal{E}_s, \mathbf{X}_s)$ is also available, which contains $n_s$ nodes with their features $ \mathbf{X}_s \in \mathbb{R}^{n_s \times d_s}$ and their normal/anomaly labels ${Y_s \in \mathbb{R}^{n_s \times 1}}$. Our task is to leverage both $\mathcal{G}_s$ and $\mathcal{G}_t$ to develop an anomaly scoring function $\phi_t$, such that:
\begin{equation}
    \phi_t(\mathcal{G}_t, v_j) \gg  \phi_t(\mathcal{G}_t, v_i) \quad \forall (v_j \in \mathcal{V}^{\mathrm{out}}_t) \land (v_i \in \mathcal{V}^{\mathrm{in}}_t),
\end{equation}
where $\mathcal{V}^{\mathrm{in}}_t$ and $\mathcal{V}^{\mathrm{out}}_t$ are respective normal and anomalous node sets, satisfying $\mathcal{V}^{\mathrm{in}}_t\cup \mathcal{V}^{\mathrm{out}}_t = \mathcal{V}_t$ and $\mathcal{V}^{\mathrm{in}}_t \cap \mathcal{V}^{\mathrm{out}}_t = \emptyset$.\par
We focus on neural-network-based anomaly scoring functions $\phi(\cdot;\Theta): \mathcal{X} \rightarrow \mathbb{R}$, which can be seen as a combination of a feature representation learner $\psi(\cdot;\Theta_f): \mathcal{X} \rightarrow \mathcal{Z}$ and a anomaly scoring function $\eta(\cdot;\Theta_g): \mathcal{Z} \rightarrow \mathbb{R}$, where $\mathcal{X}$ is the input space, $\mathcal{Z} \in \mathbb{R}^M$ is the intermediate node representation space and $\Theta = \{\Theta_f, \Theta_g\}$ are the learnable parameters of $\phi$. 
Then we aim to learn the following anomaly scoring mapping:
\begin{equation}
    \phi_t(\mathcal{G}_t, v; \Theta) = \eta_t(\psi_t(\mathcal{G}_t,v;\Theta_f^t); \Theta_g^t),
\end{equation}
with the support from $\psi_s$ and $\eta_s$ trained on the labelled source graph data. Two main challenges here include the unknown anomaly distribution in the target data, and the complex discrepancies in graph structures and semantic attribute spaces among different graphs. 

\subsection{Overview of ACT}
To address the above two challenges, we propose the approach Anomaly-aware Contrastive alignmenT (ACT). The key idea is to adapt the anomaly-discriminative knowledge from a labelled source graph to learn anomaly-informed detection models on the target graph, reducing the high detection error rates in unsupervised detection models that are lacking knowledge about the anomalies of interest. 

As illustrated in Figure \ref{fig:fw}, 
ACT learns such anomaly-informed models on the unlabelled target graph using two major components. It first performs a joint optimisation of anomaly-aware one-class domain alignment and unsupervised contrastive node representation learning on the target graph, resulting in an expressive node representation mapping $\psi_t^*$ that is domain-adapted for GAD on the target graph. 

In the second phase, ACT performs self-labelling-based deviation learning, in which an off-the-shelf anomaly detector is used on top of the domain-adapted representation space $\psi_t^*$ to identify pseudo anomalies that are subsequently employed to learn an anomaly scoring neural network on the target graph via a deviation loss. 

After domain alignment, the source-domain-based anomaly scoring network $\eta_s$ can also be used to produce the pseudo anomalies for the subsequent deviation learning, but it is generally less effective than using the off-the-shelf anomaly detector on $\psi_t$ (see Suppl. Material). Thus, the latter approach is used by default.

\subsection{Joint Learning of Contrastive Representations and Anomaly-aware Alignment}
We aim to achieve anomaly-aware one-class domain alignment in the presence of a large domain gap in graph structure, node attribute semantics, and anomaly distributions. Many UDA methods exploit pretrained representation learners or parameter sharing to initialise target node representations so that they are reasonably aligned with their corresponding source classes. However, this does not apply to graph data due to high discrepancies across different graphs.

To address this challenge, we introduce a batch-sampling-based joint learning approach to perform an optimal-transport-based domain alignment $\mathcal{L}_{\mathrm{dom}}$ with unsupervised contrastive graph learning $\mathcal{L}_{\mathrm{con}}$ by optimising the following loss function:
\begin{equation}\label{eq:alignment}
    \mathcal{L}_{\mathrm{joint}}(\mathbf{Z}_s, \mathbf{Z}_t) = \mathcal{L}_{dom}(\mathbf{Z}_s, \mathbf{Z}_t) + \mathcal{L}_{\mathrm{con}}(\mathbf{Z}_t),
\end{equation}
where $\mathbf{Z}_s= \psi_s(\mathcal{G}_s, \mathbf{B}_s;\Theta_f^s)$ and $\mathbf{Z}_t = \psi_t(\mathcal{G}_t, \mathbf{B}_t;\Theta_f^t)$ are the respective node representations of a sampled source node batch $\mathbf{B}_s$ and a target node batch $\mathbf{B}_t$. Below we introduce each term of Eq. (\ref{eq:alignment}) in detail.

\subsubsection{Unsupervised Contrastive Learning of Normal Representations of Nodes on the Target Graph}
Our unsupervised contrastive learning aims to (i) achieve initial representations of regular patterns embedded in the majority of nodes (i.e., normal representations of nodes) on the target graph and (ii) correct misalignment of node representations during the joint learning. 
To this end, we adopt a topology-based contrastive loss based on the common \textit{graph homophily phenomenon} -- similar nodes are more likely to attach to each other than dissimilar ones -- to learn the representation of target nodes. The phenomenon of homophily is assumed to be widely applied to most nodes of a graph. Thus, we use this property to define normal nodes as the ones that are consistent with their neighbourhood, and the nodes that violate the assumption are considered to be abnormal otherwise. 
Accordingly, we use this property to devise the unsupervised contrastive learning loss as:
\begin{equation}\label{eq:contrastive}
    \begin{split}
    \mathcal{L}_{\mathrm{con}}(\mathbf{Z}_t) =& - \log{(\sigma({\mathbf{Z}^{u}_{t}}^\top \mathbf{Z}^{v}_{t}))} \\
    & - Q \cdot \mathbb{E}_{v_n \sim P_n(v)} \log{(\sigma({-\mathbf{Z}^{u}_{t}}^\top \mathbf{Z}^{v_n}_{t})}),
\end{split}
\end{equation}
where $\mathbf{Z}_t = \psi_t(\mathcal{G}_t, \mathbf{B}_t;\Theta_f^t)$ are the representations of a target node batch $\mathbf{B}_t$, parameterised by $\Theta_f^t$; $\mathbf{B}_t$ consists of the target nodes $\mathbf{B}^u_t$, their positive examples $\mathbf{B}^v_t$ that occur in the first-order neighbourhood of $u$, and their negative examples  
$\mathbf{B}^{v_n}_t$ sampled from non-neighbour node set $P_n$;
$\mathbf{Z}^u_t$, $\mathbf{Z}^v_t$, and $\mathbf{Z}^{v_n}_t]$ are the representations of $\mathbf{B}_u$, $\mathbf{B}_v$ and $\mathbf{B}_{v_n}$, respectively; $\mathbf{Z_t} = [\mathbf{Z}^u_t, \mathbf{Z}^v_t, \mathbf{Z}^{v_n}_t]$ is the concatenation of the three representations. 
By minimising Eq. \ref{eq:contrastive}, the target node representation mapping $\psi_t$ is enforced to learn the regularity representations of the nodes, which can also help correct possible misalignment of the target nodes when jointly optimising with the following domain alignment.

\subsubsection{Anomaly-aware One-class Domain Alignment}
Since the target anomaly distribution can be substantially dissimilar to the source anomaly class, we propose to focus on aligning the normal class between the two domains, with the anomaly class information in the source graph to support this one-class alignment.

The choice of domain discrepancy description is crucial for the alignment. The probability-based measures and adversarial-learning-based approaches are two popular solutions \cite{uda_sur_diane}. In our case, the former approach is more appropriate than the latter one as the adversarial learning can be easily affected by the two main challenges mentioned above. The Wasserstein metric has been shown to be more promising among all probability-based discrepancy measures because it considers the underlying geometry of the probability space. It can provide reasonable measures in extreme cases, such as distributions that do not share support \cite{swd} or provide stable gradients for points that lie in low probability regions \cite{WDGRL}. Thus, we use the Wasserstein distance to measure the domain discrepancy of the normal class in the feature representation space of the two domains in an unsupervised way, while at the same time having anomaly-aware normal class representation learning in the source domain. In particular, we define the one-class alignment loss as:\par
\begin{equation}
    \mathcal{L}_{\mathrm{dom}}(\mathbf{Z}_s, \mathbf{Z}_t) = W_p(\mathbf{Z}_s, \mathbf{Z}_t),
\end{equation}
where $W_p$ is the Wasserstein distance and defined as:
\begin{equation}
    W_p(\mathbf{P}_s, \mathbf{P}_t) =  \inf_{\gamma \in \Pi} \bigg\{\bigg(\big( \mathop{\mathbb{E}}_{\mathbf{z}_s \sim \mathbf{P}_s, \mathbf{z}_t \sim \mathbf{P}_t} d(\mathbf{z}_s, \mathbf{z}_t)^p\big)^{\frac{1}{p}} \bigg)\bigg\},
\end{equation}
where $\Omega_{s}$ and $\Omega_{t}$ are two domains on a metric space $\Omega$, which are respectively related to two different probability distributions $\mathbf{P}_s$ and $\mathbf{P}_t$; $\gamma \in \pi$ is the set of all probabilistic coupling between $\Omega_{s}$ and $\Omega_{t}$; and $d(\mathbf{z}_s, \mathbf{z}_t)^p$ specifies the cost of moving any $\mathbf{z}_s \in \Omega_{s}$ to $\mathbf{z}_t \in \Omega_{t}$. We use the Sinkhorn \cite{sinkhorn} approximation of 2-Wasserstein distance for efficient estimation of the distance $d$. 

Meanwhile, to leverage the anomaly information to learn normal class representations in the source domain without enforcing any assumptions on the anomaly distribution, we use a loss function, called deviation loss \cite{devnet}. It enforces the clustering of normal nodes in the representation space w.r.t. a given prior, while making that of the anomalous nodes significantly deviate from the representations of normal nodes. Specifically, the loss adapted to our problem is given as follows:
\begin{equation}\label{eq:dev}
\begin{split}
    L(\mathbf{z}_v, \mathbf{Z}_s, \mu, \sigma) = & (1-y)\vert \mathrm{dev}(\mathbf{z}_v, \mathbf{Z}_s, \mu, \sigma)\vert + \\ & y \times \mathrm{max}(0, a - \mathrm{dev}(\mathbf{z}_v, \mathbf{Z}_s, \mu, \sigma)),
\end{split}
\end{equation}
where $\mathbf{Z}_s=\psi_s(\mathcal{G}_s,\mathbf{B}_s;\Theta_f^s)$ are the feature representations of nodes in the source graph; $y=1$ if $v$ is an anomalous node and $y=0$ otherwise; $\mathbf{z}_v \in \mathbf{Z}_s$; $a$ is a confidence interval-based margin; and $\mathrm{dev}(\mathbf{z}_v, \mathbf{Z}_s, \mu, \sigma)$ is a Z-Score-based deviation function:
\begin{equation}
    \mathrm{dev}(\mathbf{z}_v, \mathbf{Z}_s, \mu, \sigma) = \frac{\eta_s(\mathbf{z}_v, \mathbf{Z}_s;\Theta_g^s) - \mu}{\sigma},
\end{equation}
where $\mu$ and $\sigma$ are two hyperparameters from a Gaussian prior $\mathcal{N}(\mu,\sigma^2)$. Following \cite{devnet}, $\mu=0$, $\sigma=1$ and $a=5$ are used in our implementation.
Eq. (\ref{eq:dev}) is minimised via the same mini-batch gradient descent approach as in the original paper. Note that the Gaussian prior in the deviation loss is made on the normal class rather than the anomaly class, so there is no specification of the anomaly distribution.

During training, a simultaneous optimisation of $\mathcal{L}_{\mathrm{con}}(\mathbf{Z}_t)$, $\mathcal{L}_{\mathrm{dom}}(\mathbf{Z}_s, \mathbf{Z}_t)$ and $L(\mathbf{z}_v, \mathbf{Z}_s, \mu, \sigma)$ can lead to unstable performance. In our implementation, we first learn $\mathbf{Z}_s$ by minimising $L(\mathbf{z}_v, \mathbf{Z}_s, \mu, \sigma))$, and then we fix $\mathbf{Z}_s$ and perform alternating optimisation of $\mathcal{L}_{\mathrm{con}}(\mathbf{Z}_t)$ and $\mathcal{L}_{\mathrm{dom}}(\mathbf{Z}_s, \mathbf{Z}_t)$.

\subsection{Self-labelling-based Deviation Learning}
After the one-class alignment above, we obtain the domain-adapted representation space $\psi_t^*$ of the target graph and the source-domain-based anomaly scoring network $\eta_s$. Even though joint learning achieves good alignments, mismatches may still exist, which may be caused by the uncertain initial state in $\psi_t$ and the large initial discrepancy between the source and target graph distributions. Thus, directly using $\eta_s$ to perform anomaly detection on the target graph can also be unstable. To mitigate such effects, we propose the use of self-labelling-based deviation learning on the target graph. The self labelling is used to refine the learned prior knowledge of anomalies by focusing on nodes with high prediction confidence in each class to generalise the heuristics of their corresponding class distributions. Inspired by \cite{ultrahigh}, we apply {\it{Cantelli's}} Inequality based thresholding method for self labelling, which is used to obtain a set of pseudo anomalies $\mathcal{O}_{\mathrm{out}}$ via: 
\begin{equation} \label{eq:thresholding}
    \mathcal{O}_{\mathrm{out}} = \{v \vert \mathbf{s}_{\mathcal{G}_t}(v) > \text{mean}_{\mathbf{s}_{\mathcal{G}_t}} + \alpha\times  \text{std}_{\mathbf{s}_{\mathcal{G}_t}}, \forall v \in \mathcal{G}_t\},
\end{equation}
where $\mathbf{s}_{\mathcal{G}_t}=\mathcal{M}(\mathcal{G}_t,\psi_t^*)$ is a score vector that contains the anomaly scores of all nodes yielded by an off-the-shelf anomaly detector $\mathcal{M}$ on the representation space $\psi_t^*$; $\mathbf{s}_{\mathcal{G}_t}(v)$ returns the anomaly score of the node $v\in \mathcal{G}_t$; $\text{mean}_{\mathbf{s}_{\mathcal{G}_t}}$ and $\text{std}_{\mathbf{s}_{\mathcal{G}_t}}$ are the mean and standard deviation of all the scores in $\mathbf{s}_{\mathcal{G}_t}$; $\alpha > 0$ is a user-defined hyperparameter.

In addition to pseudo anomaly detection, to perform deviation learning as in Eq. \ref{eq:dev}, we also need a set of pseudo normal nodes in the target graph. Unlike pseudo anomaly identification, the identification of pseudo normal nodes is trivial, since the majority of nodes are assumed to be normal. We simply select the bottom $q$ percentile $p_{1-q}$ ranked nodes w.r.t. $\mathbf{s}_{\mathcal{G}_t}$ as the pseudo normal nodes; and the final GAD performance is insensitive to $q$. After that, we use the pseudo labelled samples to re-learn the $\psi_{t}$ by minimising the deviation loss in Eq. (\ref{eq:dev}) with $\mathbf{Z}_s$ replaced with the pseudo anomalous and normal target nodes. In doing so, it can largely reduce the effect of potential misaligned node representations in the domain alignment, since the self labelling helps effectively reduce the false positives. This optimisation accordingly produces the target-domain-based anomaly scoring network $\eta_t$, which is used together with the newly learned $\psi_{t}$ to perform anomaly detection on the target graph.


\section{Experiments}
\begin{table*}[t]
    \caption{AUC-ROC and AUC-PR (±std) comparison. 
    `N/A' indicates that CMDR (s) cannot work on datasets with different numbers of node attributes in the two domains. The boldfaced and underlined are the best and second-best results, respectively.} 
    \label{tab:main}
    \centering
    \begin{adjustbox}{width=\linewidth}
        \bgroup
        \def\arraystretch{1.1}
    \begin{tabular}{ccc|cc|cc|ccc|c|c}
    \toprule
    \multirow{2}{*}{}       & \multirow{2}{*}{\textbf{Type}}    & \multirow{2}{*}{\textbf{Method}}  & \multicolumn{8}{c|}{\textbf{CD-GAD Dataset}} & \\ \cmidrule(lr){4-5} \cmidrule(lr){6-7} \cmidrule(lr){8-10} \cmidrule(lr){11-11}
                            & &  & \textbf{RES $\rightarrow$ HTL} & \textbf{NYC $\rightarrow$ HTL} & \textbf{HTL $\rightarrow$ RES} & \textbf{NYC $\rightarrow$ RES}  & \textbf{RES $\rightarrow$ NYC} & \textbf{HTL $\rightarrow$ NYC} & \textbf{AMZ $\rightarrow$ NYC} & \textbf{NYC $\rightarrow$ AMZ} & \textbf{Average}  \\ \cmidrule[0.6pt](){1-12} 
    \multirow{11}{*}{\rotatebox[origin=c]{90}{\textbf{AUC-ROC}}}   
                            & \multirow{9}{*}{Unsup}          &G.IM + LOF       &\multicolumn{2}{c|}{\underline{0.778±0.009}} &\multicolumn{2}{c|}{\underline{0.850±0.026}} &\multicolumn{3}{c|}{0.612±0.015} &0.728±0.015  &0.742±0.016  \\
                            &                                 &G.IM + IF        &\multicolumn{2}{c|}{0.667±0.009} &\multicolumn{2}{c|}{0.843±0.017} &\multicolumn{3}{c|}{\textbf{0.844±0.007}} &0.537±0.015&0.723±0.009  \\
                            &                                 &ANOMALOUS        &\multicolumn{2}{c|}{0.186±0.002} &\multicolumn{2}{c|}{0.417±0.009} &\multicolumn{3}{c|}{0.536±0.001} &0.496±0.003  &0.409±0.004  \\
                            &                                 &DOMINANT         &\multicolumn{2}{c|}{0.694±0.000} &\multicolumn{2}{c|}{0.767±0.000} &\multicolumn{3}{c|}{0.692±0.000} &\underline{0.867±0.000}  &0.755±0.000  \\
                            &                                 &ADONE            &\multicolumn{2}{c|}{0.738±0.035} &\multicolumn{2}{c|}{0.477±0.024} &\multicolumn{3}{c|}{0.623±0.036} &0.847±0.052  &0.671±0.037  \\
                            &                                 &GAAN             &\multicolumn{2}{c|}{0.644±0.010} &\multicolumn{2}{c|}{0.668±0.041} &\multicolumn{3}{c|}{0.406±0.006} &0.861±0.015  &0.645±0.018  \\
                            &                                 &COLA             &\multicolumn{2}{c|}{0.485±0.034} &\multicolumn{2}{c|}{0.555±0.063} &\multicolumn{3}{c|}{0.811±0.006} &0.496±0.003  &0.587±0.027   \\
                                                              \cmidrule(lr){3-12}
                            & \multirow{3}{*}{\parbox{1.15cm}{\centering Cross - domain}}   &ADDA             &0.624±0.064 &0.589±0.109 &0.787±0.144 &0.726±0.345 &0.750±0.076 &0.750±0.062 &0.697±0.126 &0.640±0.051 &0.684±0.118      \\
                            &                                 &CMDR (s)         &0.690±0.009 &N/A &0.774±0.007 &N/A &N/A &N/A &0.699±0.006 &0.859±0.007 &\underline{0.756±0.007}       \\
                            &                                 &CMDR (u)         &0.699±0.009 &0.707±0.009 &0.763±0.024 &0.780±0.017 &0.694±0.006 &0.693±0.002 &0.695±0.001 &0.848±0.009 &0.751±0.012    \\ \cmidrule(lr){3-12}
                            & Ours                            &ACT              &\textbf{0.804±0.006} &\textbf{0.792±0.018} &\textbf{0.892±0.015} &\textbf{0.948±0.014} &\underline{0.831±0.005} &\underline{0.830±0.005} &\underline{0.830±0.002} &\textbf{0.925±0.004} &\textbf{0.868±0.009}   \\ \cmidrule(r){2-12}\morecmidrules\cmidrule(r){2-12} 
    \multirow{11}{*}{\rotatebox[origin=c]{90}{\textbf{AUC-PR}}}  
                            & \multirow{9}{*}{Unsup}          &G.IM + LOF       &\multicolumn{2}{c|}{\underline{0.247±0.016}} &\multicolumn{2}{c|}{0.269±0.036} &\multicolumn{3}{c|}{0.133±0.006} &0.097±0.008  &0.186±0.017  \\
                            &                                 &G.IM + IF        &\multicolumn{2}{c|}{0.194±0.013} &\multicolumn{2}{c|}{\underline{0.296±0.031}} &\multicolumn{3}{c|}{\textbf{0.366+0.014}} &0.042±0.003  &\underline{0.226±0.009}      \\
                            &                                 &ANOMALOUS        &\multicolumn{2}{c|}{0.053±0.000} &\multicolumn{2}{c|}{0.040±0.001} &\multicolumn{3}{c|}{0.091±0.001} &0.037±0.000  &0.055±0.001      \\
                            &                                 &DOMINANT         &\multicolumn{2}{c|}{0.216±0.000} &\multicolumn{2}{c|}{0.264±0.000} &\multicolumn{3}{c|}{0.145±0.000} &0.252±0.000  &0.219±0.000      \\
                            &                                 &ADONE            &\multicolumn{2}{c|}{0.244±0.029} &\multicolumn{2}{c|}{0.183±0.031} &\multicolumn{3}{c|}{0.155±0.029} &\underline{0.259±0.076}  &0.210±0.041     \\
                            &                                 &GAAN             &\multicolumn{2}{c|}{0.152±0.006} &\multicolumn{2}{c|}{0.089±0.017} &\multicolumn{3}{c|}{0.039±0.001} &0.203±0.035  &0.121±0.015      \\
                            &                                 &COLA             &\multicolumn{2}{c|}{0.082±0.009} &\multicolumn{2}{c|}{0.109±0.011} &\multicolumn{3}{c|}{0.128±0.003} &0.037±0.000  &0.089±0.006   \\
                                                              \cmidrule(lr){3-12}
                            & \multirow{3}{*}{\parbox{1.15cm}{\centering Cross - domain}}    &ADDA             &0.227±0.028 &0.171±0.062 &0.260±0.126 &0.254±0.140 &\underline{0.254±0.140} &0.181±0.021 &0.239±0.110 &0.051±0.002 &0.177±0.057       \\
                            &                                 &CMDR (s)         &0.210±0.007 &N/A &0.268±0.006 &N/A &N/A &N/A &0.145±0.001 &0.242±0.019 &0.216±0.008        \\
                            &                                 &CMDR (u)        &0.216±0.008 &0.207±0.009 &0.253±0.025 &0.267±0.015 &0.144±0.003 &0.144±0.002 &0.145±0.001 &0.220±0.024 &0.209±0.014    \\ \cmidrule(lr){3-12}
                            & Ours                            &ACT              &\textbf{0.287±0.006} &\textbf{0.284±0.010} &\textbf{0.330±0.018} &\textbf{0.477±0.065} &0.249±0.012 &\underline{0.241±0.009} &\underline{0.243±0.003} &\textbf{0.497±0.020} &\textbf{0.358±0.002}    \\ 
                            \bottomrule \\
    \end{tabular}
    \egroup
    \end{adjustbox}
\end{table*}

\subsection{Datasets}
Eight CD-GAD settings based on four real-world GAD datasets, including \textit{YelpHotel} (HTL), \textit{YelpRes} (RES), \textit{YelpNYC} (NYC) and \textit{Amazon} (AMZ)\footnote{Statistics of each dataset are given in Suppl. Material}, are created as follows, with each setting having two related datasets as the source and target domains. 

\textit{YelpHotel} (HTL) $\rightleftharpoons$ \textit{YelpRes} (RES). These two datasets are Yelp online review graphs in the Chicago area for accommodation and dining businesses. A node represents a reviewer and an edge indicates two reviewers have reviewed the same business. Reviewers with filtered reviews by Yelp anti-fraud filters are regarded as anomalies. Each of the datasets can serve as either source or target domain. The primary domain shift here is the course of business. \par
\textit{YelpNYC} (NYC)$\rightleftharpoons$ \textit{Amazon} (AMZ). These are also review graphs.
YelpNYC is collected from New York City for dining businesses, while Amazon is for E-commerce reviews. Anomalies are users with multiple reviews identified using crowd-sourcing efforts. The domain gap here is greater than HTL $\rightleftharpoons$ RES as these two datasets are less co-related. \par
\textit{YelpRes} (RES) $\rightleftharpoons$ \textit{YelpNYC} (NYC). The primary domain shift here is geographical location, as both graphs are for dining business reviews. This pair presents additional significant challenges due to their heterogeneous feature spaces and a large difference in graph size.\par
\textit{YelpHotel} (HTL) $\rightleftharpoons$ \textit{YelpNYC} (NYC). It is similar to RES $\rightleftharpoons$ NYC, however, with more substantial domain gaps in not only geographical locations
but also their business types (dining venues vs. accommodation).

\subsection{Competing Methods and Evaluation Metrics}
We consider 10 SOTA competing methods from two related lines of research: unsupervised GAD and CD methods.  
Two unsupervised GAD methods are based on the combination of LOF \cite{lof} and iForest (IF) \cite{if} and node embedding via Deep Graph Infomax \cite{im}. Further, we also include five recent unsupervised GAD methods
: ANOMALOUS \cite{anomalous}, DOMINANT \cite{domi}, AdONE \cite{adone}, GGAN \cite{ggan} and COLA \cite{cola}. They are included to examine whether ACT can benefit from the source domain information for unsupervised GAD on the target domain. For CD methods, we choose COMMANDER \cite{cmd} (CMDR for short)
and ADDA -- a popular general domain adaptation method \cite{uda_sur_diane}. As the original CMDR, termed CMDR (s), adopts a shared representation learner for both domains, we derive a variant of CMDR, termed CMDR (u), that can work in two domains with different feature spaces by learning separate graph representation learners for each domain.

We employ two popular, complementary performance metrics for AD, the Area Under Receiver Operating Characteristic Curve (AUC-ROC) and the Area Under Precision-Recall Curve (AUC-PR), which are holistic metrics that quantify the performance of an AD model across a wide range of decision thresholds. Larger AUC-ROC (or AUC-PR) indicates better performance.

\subsection{Implementation Details}
Our model ACT is implemented with a three-layer GraphSAGE \cite{sage} within which 256 and 64 hidden dimensions are chosen for $\psi_s$ and $\psi_t$ respectively. 
The source model is trained for 50 epochs using a learning rate of $10^{-3}$. The domain alignment is performed for 50 epochs using a learning rate of  $10^{-4}$. The same learning rate is also used in self-labelling-based deviation learning, wherein IF is used as the off-the-shelf detector $\mathcal{M}$.
The optimisation is done in mini-batches of 128 target (centre) nodes using the ADAM optimiser \cite{adam} . We use the sample size of 25 and 10 for the two hidden layers during message passing. In self labelling, $\alpha=2.5$ and $q=25$ are used by default.
These neural network settings and training methods are used throughout all the settings of our experiments. All the results are averaged over five independent runs using random seeds. The model settings and training of the competing methods are based on default/recommended choices of their authors.

\subsection{Detection Performance on Real-world Datasets}
We compare ACT with 10 SOTA competing methods on eight real-world CD settings, with the results shown in Table~\ref{tab:main}, where `A $\rightarrow$ B' represents 
the use of a source dataset A for GAD on a target dataset B; and unsupervised anomaly detectors use only the target data. 

\noindent\textbf{Overall Performance} ACT performs stably across all eight settings and substantially outperforms all competing methods by at least 11\% and 13\% in average AUC-ROC and AUC-PR, respectively. In particular, benefiting from the anomaly-aware alignment and self-labelling-based deviation learning, ACT demonstrates consistent superiority over the competing CD methods on all eight datasets. Unsupervised detectors work well only on very selective datasets where their definition of anomaly fits well with the underlying anomaly distribution, e.g., the method IF on NYC, and they become unstable and ineffective otherwise. By contrast, ACT learns anomaly-informed models with the relevant anomaly supervision from the source data, and thus, it can perform stably and work well across the datasets. 

\noindent\textbf{Semantic Domain Gap} For CD-GAD, 
using different source graphs results in similar performance in most cases. However, in some cases, one source can be more informative than the others, e.g., the results of ACT on NYC $\rightarrow$ RES vs. HTL $\rightarrow$ RES, indicating a closer domain gap between NYC and RES than that between HTL and RES.

\noindent\textbf{Heterogeneous Structure/Attribute Inputs} ACT can effectively handle scenarios where the source and the target have a large difference in graph structure and/or node attribute dimension, such as NYC $\rightarrow$ HTL and NYC $\rightarrow$ RES.
By contrast, ADDA and CMDR (u) fail to work effectively in such cases (CMDR (s) is inapplicable as it requires a shared feature learner on the two domains).

\begin{figure}[t]
    \centering
    \includegraphics[width=0.9\linewidth]{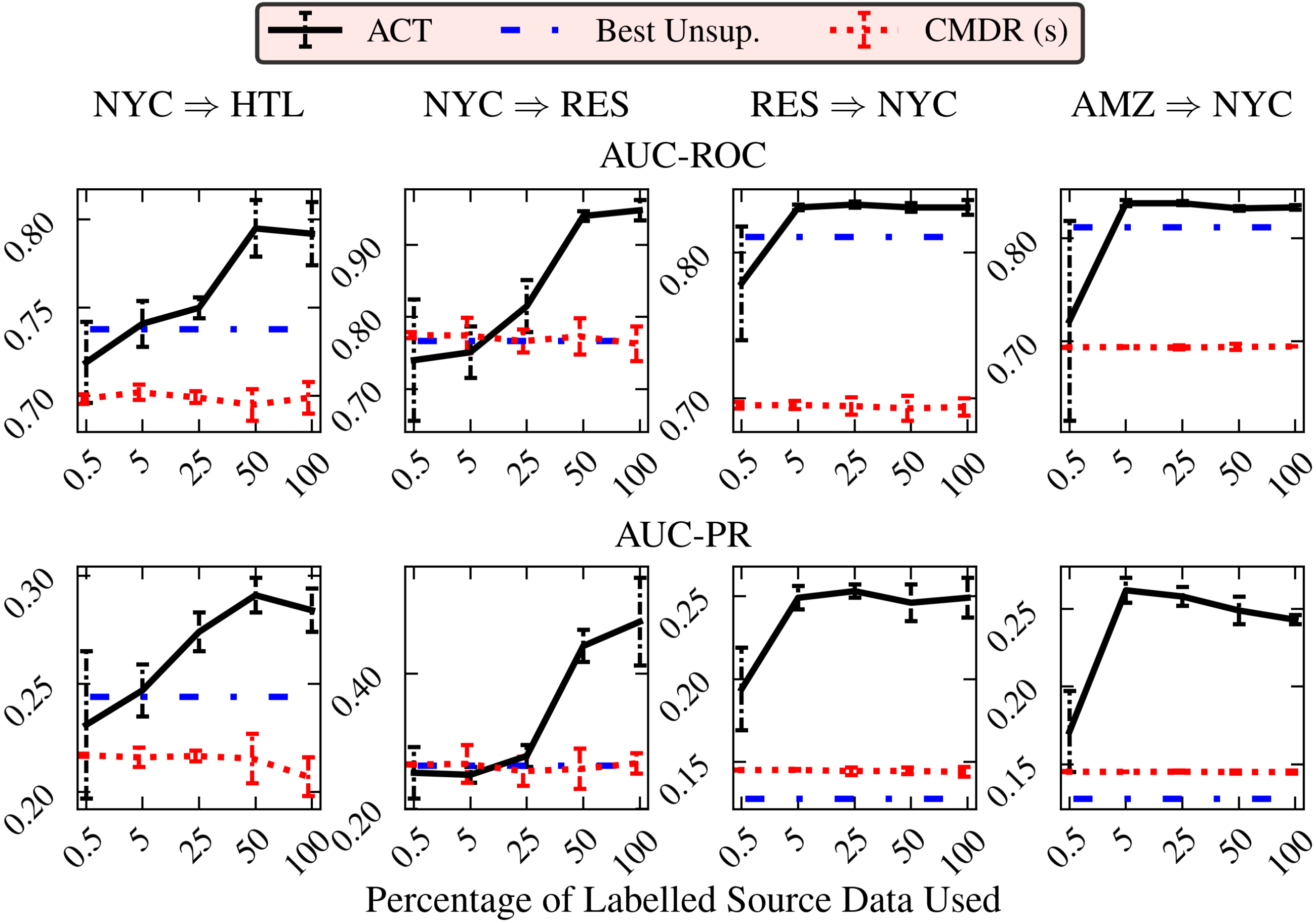}
    \caption{Efficiency of ACT utilising labelled source data, with best CD/unsupervised models as baselines.}
    \label{fig:de}
\end{figure}

\subsection{Effectiveness of Utilising Source Domain Data}

This subsection provides an in-depth empirical investigation of the importance of source data to CD-GAD by answering two key questions below.

\noindent\textbf{How much source domain data is required by ACT to outperform SOTA unsupervised detectors?}
To answer this question, we evaluate the performance of ACT on four representative datasets of different data complexities using five percentages of labelled source nodes: 0.5\%, 5\%, 25\%, 50\% and 100\% (the rest of the nodes are treated as unlabelled data during training). The results are illustrated in Figure \ref{fig:de}, with CMDR (s) and the best unsupervised result per dataset as baselines. It is impressive that even when a very small percentage (e.g., 0.5\% or 5\%) of labelled source data is used, ACT can perform better, or on par with, these strong baselines, demonstrating strong capability in unleashing the relevant information hidden in the source data. 
This capability is further verified by the increasing AUC-ROC and AUC-PR of ACT 
when the amount of source data used increases. Nevertheless, caution is required when the labelled source data is too small (e.g., 0.5\% labelled source data corresponds to 21 nodes to 105 nodes for the four datasets), since ACT can perform unstably in such cases.  

\begin{table}[]
    \caption{Self-labelling deviation learning on our ACT-based domain-adapted feature space vs. the original feature space. METHOD$^*$ means the use of METHOD to perform self labelling in the original feature space and then performs exactly the same deviation learning as in ACT using Eq. (\ref{eq:dev}). }
    \label{tab:plus_devnet}
        \centering
        \begin{adjustbox}{width=\columnwidth}
        \bgroup
        \def\arraystretch{1.1}
        \begin{tabular}{ccccccc}
        \toprule
                             &                             & \textbf{RES} $\rightarrow$ \textbf{HTL} & \textbf{HTL} $\rightarrow$ \textbf{RES} & \textbf{NYC} $\rightarrow$ \textbf{RES} &\textbf{NYC} $\rightarrow$ \textbf{AMZ} \\ \cmidrule[0.6pt](){1-6} 
                             \multirow{6}{*}{\rotatebox[origin=c]{90}{\textbf{AUC-ROC}}}   
                             & ANOMALOUS$^*$                    & 0.434±0.025           &\multicolumn{2}{c}{0.594±0.051}         & 0.434±0.006           \\
                             & DOMINANT$^*$                     &0.737±0.005            &\multicolumn{2}{c}{0.914±0.008}         &0.912±0.002            \\
                             & ADONE$^*$                        &0.674±0.059            &\multicolumn{2}{c}{0.825±0.057}         &0.775±0.089            \\
                             & GGAN$^*$                         &0.664±0.010            &\multicolumn{2}{c}{0.851±0.014}         &0.855±0.027            \\
                             & COLA$^*$                         &0.522±0.028            &\multicolumn{2}{c}{0.683±0.070}         &0.730±0.012            \\\cmidrule(r){2-6}
                             & Ours                         &\textbf{0.804±0.006}           &0.892±0.015        &\textbf{0.948±0.014}         &\textbf{0.925±0.004}            \\ \cmidrule(r){2-6}\morecmidrules\cmidrule(r){2-6} 
                             \multirow{6}{*}{\rotatebox[origin=c]{90}{\textbf{AUC-PR}}}   
                             & ANOMALOUS$^*$    & 0.098±0.009 &\multicolumn{2}{c}{0.126±0.015} &0.038±0.002        \\
                             & DOMINANT$^*$     &0.277±0.004            &\multicolumn{2}{c}{0.366±0.013}           &0.383±0.006            \\
                             & ADONE$^*$     &0.243±0.034            &\multicolumn{2}{c}{0.288±0.026}           &0.195±0.140            \\
                             & GGAN$^*$      &0.247±0.011            &\multicolumn{2}{c}{0.296±0.016}           &0.297±0.051            \\
                             & COLA$^*$      &0.163±0.043            &\multicolumn{2}{c}{0.224±0.017}           &0.096±0.010            \\\cmidrule(r){2-6}
                             & Ours      &\textbf{0.287±0.006}    &0.330±0.018        &\textbf{0.477±0.065}           &\textbf{0.497±0.020}            \\ \bottomrule
    \end{tabular}
    \egroup
    \end{adjustbox}
\end{table}

In addition to the domain alignment component, another major factor in the superior performance of ACT here is the self-labelling deviation learning (see our Ablation Study). Therefore, the second question below is investigated.

\noindent\textbf{Can we just perform self-labelling deviation learning on the target domain directly, without using any source domain data?} The answer is clearly negative. This can be observed by our empirical results in Table \ref{tab:plus_devnet}, where ACT is compared with five deviation-learning-enhanced unsupervised competing methods on four representative settings that cover adaptations between graphs of similar/different sizes and attributes.
The results show that although self-labelling deviation learning helps achieve performance improvements on several datasets compared to the results of the original five unsupervised methods in Table \ref{tab:main}, ACT still outperforms these five enhanced baselines by substantial margins in both AUC-ROC and AUC-PR. These results indicate that there is crucial anomaly knowledge adapted from the source data in the domain alignment stage in ACT; such knowledge cannot be obtained by working on only the target data. 

\subsection{Ablation Study}

\noindent\textbf{Joint Contrastive Graph Representation Learning and Anomaly-aware Alignment}
We first evaluate the importance of synthesising contrastive learning on the target graph and anomaly-aware domain alignment ($\mathcal{L}_{\mathrm{joint}}$) in ACT, compared to the use of the individual contrastive learning ($\mathcal{L}_{\mathrm{con}}$) or anomaly-aware alignment ($\mathcal{L}_{\mathrm{dom}}$). 
The results are reported in Table~\ref{tab:joint}, which shows that the joint learning enables significantly better adaptation of anomaly knowledge in the source domain to the target domain, substantially outperforming the use of $\mathcal{L}_{\mathrm{con}}$ or $\mathcal{L}_{\mathrm{dom}}$
across the eight settings. $\mathcal{L}_{\mathrm{joint}}$ outperforms the two ACT variants by at least 12\% and 14\% in average AUC-ROC and AUC-PR respectively.
The joint learning is advantageous because the contrastive learning models the regular patterns of the nodes in the target graph (i.e., learning the representations of normal nodes),
while the anomaly-aware domain alignment allows the use of labelled anomaly and normal nodes in the source data to improve the normal representations in the target data. Optimising these two objectives independently fails to work effectively due to their strong reliance on each other.

\begin{table}[h]
    \caption{Anomaly-aware contrastive alignment vs. separate contrastive learning/anomaly-aware alignment.}
    \label{tab:joint}
    \centering
    \begin{adjustbox}{width=0.8\columnwidth}
        \bgroup
        \def\arraystretch{1.0}
    \begin{tabular}{ccccccc}
    \toprule
    \textbf{}                      &\multicolumn{3}{c}{\textbf{AUC-ROC}}      &\multicolumn{3}{c}{\textbf{AUC-PR}}        \\ \cmidrule(lr){2-4} \cmidrule(lr){5-7}
    \textbf{}                      &$\mathcal{L}_{\mathrm{con}}$ & $\mathcal{L}_{\mathrm{dom}}$ &$\mathcal{L}_{\mathrm{joint}}$ &$\mathcal{L}_{\mathrm{con}}$ & $\mathcal{L}_{\mathrm{dom}}$ &$\mathcal{L}_{\mathrm{joint}}$    \\ \midrule
    \textbf{RES $\rightarrow$ HTL} &0.485  &\textbf{0.610}  &0.608 &0.099 &0.216 &\textbf{0.216}      \\
    \textbf{NYC $\rightarrow$ HTL} &0.534  &0.609  &\textbf{0.682} &0.125 &0.211 &\textbf{0.246}     \\
    \textbf{HTL $\rightarrow$ RES} &0.552  &0.862  &\textbf{0.880} &0.069 &\textbf{0.308} &0.296      \\
    \textbf{NYC $\rightarrow$ RES} &0.596  &0.662  &\textbf{0.961} &0.316 &0.160 &\textbf{0.444}      \\
    \textbf{HTL $\rightarrow$ NYC} &0.615  &0.671  &\textbf{0.773} &\textbf{0.179} &0.145 &0.171      \\
    \textbf{RES $\rightarrow$ NYC} &0.437  &0.675  &\textbf{0.753} &0.096 &0.143 &\textbf{0.163}      \\
    \textbf{AMZ $\rightarrow$ NYC} &0.578  &0.578  &\textbf{0.617} &0.101 &0.134 &\textbf{0.154}      \\
    \textbf{NYC $\rightarrow$ AMZ} &0.389  &0.640  &\textbf{0.880} &0.033 &0.078 &\textbf{0.587}      \\ \cmidrule(lr){1-1} \cmidrule(lr){2-4} \cmidrule(lr){5-7}
    \textbf{Average}               &0.523  &0.663  &\textbf{0.790} &0.127 &0.179 &\textbf{0.338}      \\ \bottomrule \\
    \end{tabular}
    \egroup
    \end{adjustbox}
\end{table}

\noindent\textbf{Self-labelling-based Deviation Learning}
We then evaluate the importance of the self-labelling-based deviation learning component in ACT, with two variants of ACT, $\eta_{s}$ and ACT-IF. $\eta_{s}$ directly uses the source-domain-based anomaly detector $\eta_{s}$, while ACT-IF uses IF on the domain-adapted feature representation space of the target data to detect anomalies; both of which are done after the anomaly-aware contrastive alignment, but they do not involve the deviation learning.

Table~\ref{tab:pseudo_ablation} shows the comparison results, from which we can observe that the self-labelling-based deviation learning component in ACT largely outperforms $\eta_{s}$ and ACT-IF,
achieving average improvement by at least 8\% in AUC-ROC and 2\% in AUC-PR. The improvement can be attributed to the capability of the self labelling in identifying true anomalies in the target data with high confidence predictions, which enhances the representation learning of the normal and anomalous nodes in the subsequent deviation learning. The large improvement in AUC-ROC and relatively small improvement in AUC-PR indicate that ACT is more effective in reducing false positives than increasing true positives.

\begin{table}[h]
    \caption{Self-labelling deviation learning in ACT vs. domain-adapted anomaly detector $\eta_s$ and unsupervised detector IF on the adapted target feature space.}
    \label{tab:pseudo_ablation}
    \centering
    \begin{adjustbox}{width=0.8\columnwidth}
        \bgroup
        \def\arraystretch{1.0}
        \begin{tabular}{ccccccc}
            \toprule
            \textbf{}                      &\multicolumn{3}{c}{\textbf{AUC-ROC}}      &\multicolumn{3}{c}{\textbf{AUC-PR}}        \\ \cmidrule(lr){2-4} \cmidrule(lr){5-7}
            \textbf{}                      &$\eta_{s}$ &ACT-IF &ACT &$\eta_{s}$ &ACT-IF &ACT    \\ \midrule
            \textbf{RES $\rightarrow$ HTL}  &0.608 &0.740  &\textbf{0.804}  &0.216   &0.261  &\textbf{0.287}      \\
            \textbf{NYC $\rightarrow$ HTL}  &0.682 &0.744  &\textbf{0.792}  &0.246   &0.257  &\textbf{0.284}    \\
            \textbf{HTL $\rightarrow$ RES}  &0.880 &0.843  &\textbf{0.892}  &0.296   &0.262  &\textbf{0.330}    \\
            \textbf{NYC $\rightarrow$ RES}  &\textbf{0.961} &0.955  &0.948  &0.444   &0.427  &\textbf{0.477}    \\
            \textbf{HTL $\rightarrow$ NYC}  &0.773 &0.781  &\textbf{0.831}  &0.171   &0.166  &\textbf{0.249}    \\
            \textbf{RES $\rightarrow$ NYC}  &0.753 &0.748  &\textbf{0.830}  &0.163   &0.158  &\textbf{0.241}    \\
            \textbf{AMZ $\rightarrow$ NYC}  &0.617 &0.792  &\textbf{0.830}  &0.154   &0.191  &\textbf{0.243}    \\
            \textbf{NYC $\rightarrow$ AMZ}  &0.880 &0.821  &\textbf{0.925}  &\textbf{0.587}  &0.556  &0.497    \\ \cmidrule(lr){1-1} \cmidrule(lr){2-4} \cmidrule(lr){5-7}
            \textbf{Average}                &0.790 &0.785  &\textbf{0.868}  &0.338   &0.333  &\textbf{0.358}    \\ \bottomrule \\
    \end{tabular}
    \egroup
    \end{adjustbox}
\end{table}

\subsection{Anomaly Thresholding Sensitivity}
\begin{figure}[h]
    \centering
    \includegraphics[width=0.9\linewidth]{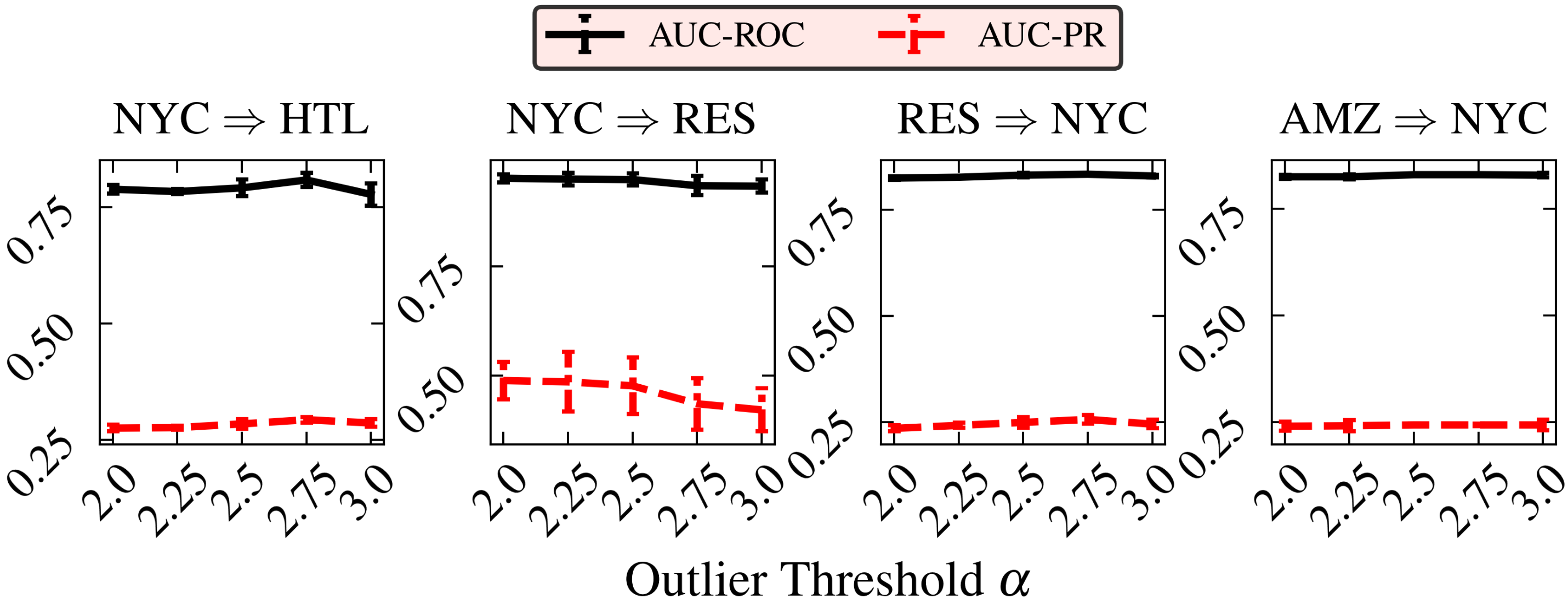}
    \caption{Sensitivity test results w.r.t. $\alpha$.}
    \label{fig:ps}
\end{figure}

This section studies the sensitivity of ACT w.r.t. the anomaly thresholding hyperparameter $\alpha$ in Eq. (\ref{eq:thresholding}), which determines the characteristics of the pseudo anomalies (e.g., the number and quality). The results with varying $\alpha$ settings are reported in Figure \ref{fig:ps}. In general, ACT maintains stable performance across the value range of $\alpha$ in $[2.0, 3.0]$, suggesting its good stability on datasets with different characteristics.

\section{Conclusion}
In this paper, we present Anomaly-aware Contrastive alignmenT (ACT) for CD-GAD, which connects an optimal-transport-based discrepancy measure and graph-structure-based contrastive loss to leverage prior AD knowledge from a source graph as a joint learning scheme. The resulting model achieves anomaly-aware one-class alignment under severe data imbalance and different source and target distributions. A self-labelling approach to deviation learning is further proposed to refine the learned source of AD knowledge. These two components result in significant GAD improvement on various real-world cross domains. In our future work, we plan to explore the use of multiple source graphs under the ACT framework.

\section{Acknowledgements}
This work was supported in part by the MASSIVE HPC facility (www.massive.org.au),
The University of Melbourne’s Research Computing Services and the Petascale Campus Initiative. Guansong Pang is supported in part by the Singapore Ministry of Education
(MoE) Academic Research Fund (AcRF) Tier 1 grant (21SISSMU031).

\bibliography{ref}

\appendix
\section{Supplementary Material for ``Cross-Domain Graph Anomaly Detection \\ via Anomaly-aware Contrastive Alignment"}

\subsection{Dataset Details}

\subsubsection{Dataset Statistics}
Table \ref{tab:dset} shows the statistics of the four datasets used, which are based on the datasets after a node down-sampling (see the subsection below) and the removal of isolated nodes.
\begin{table}[H]
    \centering
    \caption{A summary of dataset statistics}
    \label{tab:dset}
    \begin{adjustbox}{width=0.9\columnwidth}
    \begin{tabular}{cccccc}
    \toprule
                        & \textbf{\# Dim.} & \textbf{\# Nodes} & \textbf{Avg Deg.} & \textbf{\# Anomalies} & \textbf{\# Ratio} \\ \cmidrule(){1-6}
    \textbf{YelpRes}    &8000       &5012       &41.79       &250    &0.0499      \\
    \textbf{YelpHotel}  &8000       &4322       &23.55       &250    &0.0578      \\
    \textbf{YelpNYC}    &10000      &21040      &78.81              &1000   &0.0475      \\
    \textbf{Amazon}     &10000      &18601      &28.30          &726    &0.0390      \\
    \bottomrule    \\         
    \end{tabular}
    \end{adjustbox}
\end{table}
\subsubsection{Dataset Source and Pre-processing}
The four real-world datasets used for empirical evaluation are requested from the authors of \cite{cmd}, which are the processed versions of the \textit{Yelp} \cite{yelp_leman} datasets and the \textit{Amazon} \cite{kaghazgaran2018combating, anton_sigir} dataset. The datasets would be released upon the approval of \citet{cmd}.

In terms of pre-processing, following \cite{sage}, we down-sample the edges in all graphs to reduce the heavy-tailed nature of degree distribution such that any node has at most 128 edges, i.e., random edge sampling is applied to any nodes that has more than 128 edges.

\subsection{Algorithm Implementations}
\subsubsection{Implementaiton of the Competing Methods}
For the method G.IM + LOF and G.IM + IF, we use the off-the-shelf implementation of Deep Graph Infomax from the PyTorch Geometric Library in combination with the LOF and the IF implementations from the Scikit-learn library. 
The other five SOTA unsupervised methods are taken from the PYGOD library, with the recommended model settings in the original papers. For ADDA, we use the same representation learner as ACT and leave the other components identical to \cite{adda}. For CMDR, we implement the model based on the paper due to the unavailability of the official implementation. 

\subsubsection{Optimisation of the the Competing Methods}
We train the Deep Graph Infomax for 50 epochs at a learning rate of $1.0 \times 10^{-4}$. We use the default settings of the LOF and IF in the Scikit-learn Library. We use the authors' recommended settings for the SOTA unsupervised baselines. ADDA are trained for 200 epochs using the same setting stated in its original paper. CMDR variants are trained as described in their original paper.

\subsubsection{Libraries and Their Versions}
A list of key libraries and their versions used in our implementation is provided as follows.

\begin{itemize}
    \item python==3.8.12
    \item pytorch==1.8.0
    \item pytorch geometric==2.0.1
    \item numpy==1.21.2
    \item scipy==1.7.1
    \item scikit-learn==1.0.1
    \item cudatoolkit==11.1.1
    \item geomloss==0.2.4
\end{itemize}
\subsubsection{Hardware Environment}
Our main experiments are performed using a single NVIDIA A40 GPU with a Intel(R) Xeon(R) Gold 5320 CPU. The CUDA driver installed on the platform is 470.129.06.

\begin{figure*}[t]
    \centering
    \includegraphics[width=0.9\linewidth]{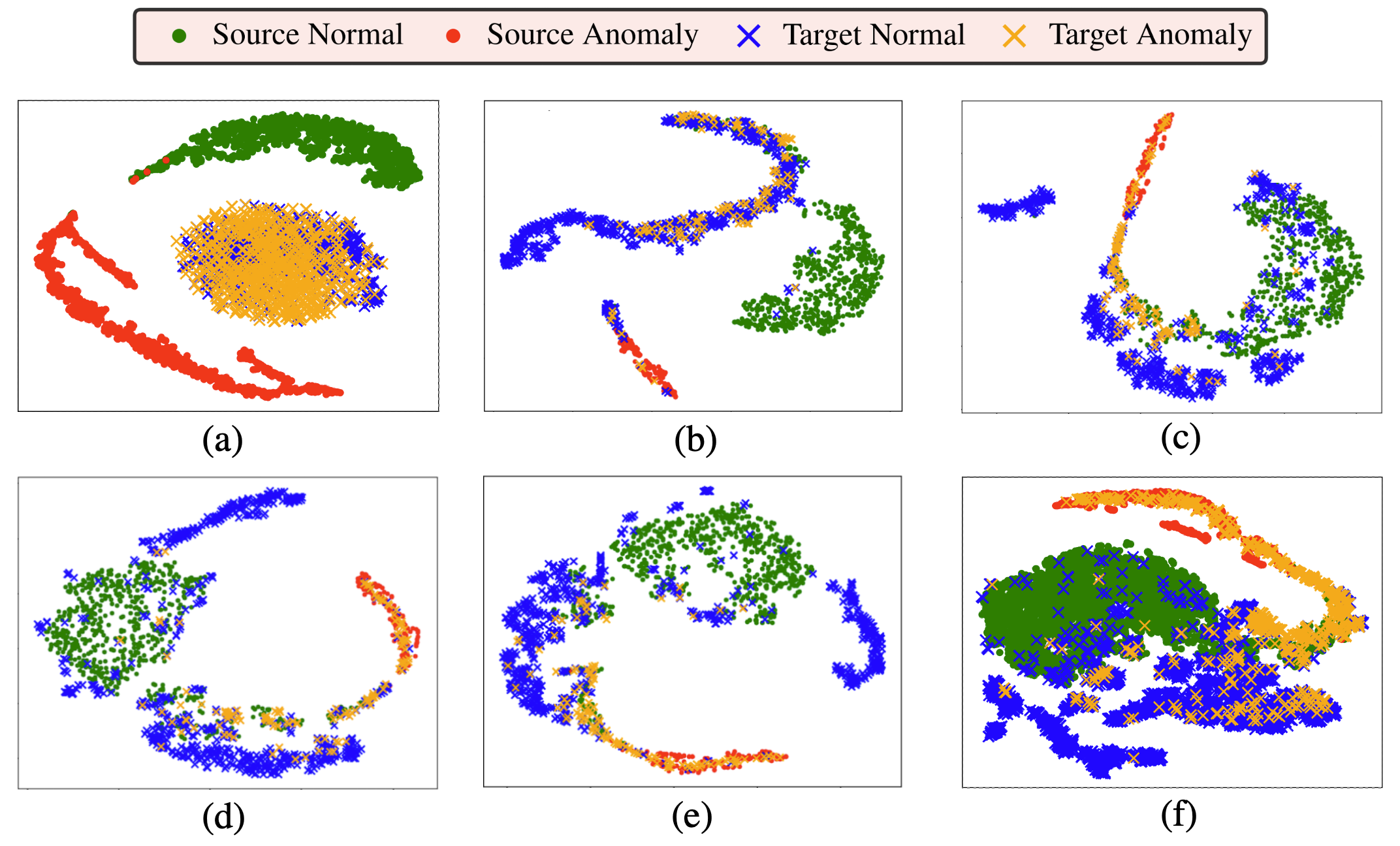}
    \caption{A visualisation of the node representation space during the ACT’s joint learning. (a): the initial state. (f): the final state. (b) $\rightarrow$ (e): the intermediate states. Note, the intermediate states are plotted using a reduced number of normal samples to allow clearer visualisation.}
    \label{fig:vis1}
\end{figure*}

\subsection{Additional Experimental Results}
\subsubsection{The Choice of the Self-labelling Method}
Recall that the second phase of ACT involves generating pseudo samples for the subsequent deviation learning. Other than ACT's default choice IF, $\eta_s$ can potentially be applied to derive the score vector $\mathcal{S}_{\mathcal{G}_t}$ on the target graph for the pseudo selection.
Here, while keeping the other components of ACT unchanged, we report the GAD performance of using $\eta_s$ for self-labelling and discuss why IF is more appropriate. \par
The results with using $\eta_{s}$ for self labelling are reported in Table \ref{tab:labeller}. The results are obtained using exactly the same model training settings as ACT.
Consistent with our claim in Section 3 in the main text, ACT yields better overall performance on most datasets than using $\eta_{s}$. This is mainly because that there can be still some domain gaps after our domain alignment, thus, there is no guarantee that the learned target node representations always fall within the correct region of the decision boundary. As a result, directly using $\eta_{s}$ on the target graph can fail to work properly in the cases where the domain gap is not small enough.
On the other hand, IF treats nodes that deviate significantly from the majority as pseudo anomalies and those located in high-density areas as normal nodes regardless of the learned decision boundary of $\eta_s$. This is in line with the characteristics of the learned representation space by the joint learning objectives in our approach ACT, resulting in better self-labelling performance and subsequently better detection performance on the target graph.
\begin{table}[h]
    \caption{Results of ACT with self-labelling using $\mathcal{\eta}_{s}$ vs IF}
    \label{tab:labeller}
\centering
 \begin{adjustbox}{width=\linewidth}
\begin{tabular}{ccccc}
\toprule
                            & \multicolumn{2}{c}{\textbf{AUROC}} & \multicolumn{2}{c}{\textbf{AUPR}} \\ \cmidrule(lr){2-3} \cmidrule(lr){4-5}
                            &$\eta_s$   &IF     &$\eta_s$    &IF    \\ \midrule 
\textbf{RES $\rightarrow$ HTL}                    &0.733±0.015             &0.804±0.006        &0.252±0.007            &0.287±0.006             \\
\textbf{NYC $\rightarrow$ HTL}                    &0.763±0.015            &0.792±0.018         &0.271±0.009             &0.284±0.010             \\
\textbf{HTL} $\rightarrow$ \textbf{RES}                    &0.842±0.050            &0.892±0.015         &0.299±0.028           &0.330±0.018             \\
\textbf{NYC} $\rightarrow$ \textbf{RES}                    &0.906±0.050            &0.948±0.014         &0.384±0.058          &0.477±0.065             \\
\textbf{RES} $\rightarrow$ \textbf{NYC}                    &0.828±0.006            &0.831±0.005         &0.238±0.016          &0.249±0.012             \\
\textbf{HTL} $\rightarrow$ \textbf{NYC}                    &0.829±0.005            &0.830±0.005         &0.245±0.012         &0.241±0.009             \\
\textbf{AMZ} $\rightarrow$ \textbf{NYC}                    &0.823±0.010            &0.830±0.002         &0.225±0.016          &0.243±0.003            \\
\textbf{NYC} $\rightarrow$ \textbf{AMZ}                    &0.930±0.008            &0.925±0.004         &0.515±0.014           &0.497±0.020            \\ \cmidrule(lr){1-5}
\multicolumn{1}{c}{\textbf{Average}}             &0.845±0.024            &0.868±0.009          &0.339±0.022            &0.358±0.002\\ \bottomrule
\end{tabular}
\end{adjustbox}
\end{table}

\subsection{Detailed Visualisation of ACT's Domain Alignment}
Figure \ref{fig:vis1} in the main text provides an example of the node embeddings of the source and target graphs before and after our anomaly-aware one-class alignment. Here we provide a more detailed visualisation of the domain alignment process in ACT on the same dataset NYC $\rightarrow$ AMZ in Figure \ref{fig:vis1}, in which each figure corresponds to the embeddings of a model checkpoint taken at a specific training epoch. The progressive results from subfigure (a) to (f) show that the embeddings of nodes, especially the normal nodes, in the target graph are continuously adapted to that of the source graph.

\subsection{The Off-the-shelf Anomaly Detector $\mathcal{M}$}
The off-the-shelf anomaly detector we employ is Isolation Forest (IF) \cite{if}, a recursive partition-based algorithm that measures the normality of observations by the number of random feature splits to subdivide their corresponding regions such that the number of observations in each region is no more than a predefined maximum. In simple words, observations that take noticeably fewer splits to isolate would locate in regions that have fewer observations compared to the majority and, thus, are more likely to be anomalies. The intuition of IF is in line with the learned representation space of ACT, where the anomaly-aware one-class alignment is achieved between the source and target normal classes and anomalies are pushed away from the normal (majority) region. 

\subsection{Pseudocode for ACT}
Algorithm \ref{alg:joint} describes the joint learning of ACT, where $\mathbf{Z}_s = \psi_s(\mathcal{G}_s, \mathbf{B}_s)$ and $\mathbf{Z}_t = \psi_t(\mathcal{G}_t, \mathbf{B}_t)$ are the representations of a source batch $\mathbf{B}_t$ and target batch $\mathbf{B}_t$, respectively. $\mathbf{B}_t$ consists of the target nodes $\mathbf{B}^u_t$, their positive examples $\mathbf{B}^v_t$ that co-occurs near each $u$ and their negative examples (Q for each $u$, $Q \cdot \vert \mathbf{B}^u_t \vert$ in total) $\mathbf{B}^{v_n}_t$ from the negative sampling distribution $P_n$. $\mathbf{Z_t} = [\mathbf{Z}^u_t, \mathbf{Z}^v_t, \mathbf{Z}^{v_n}_t]$ concatenates the representations of $\mathbf{B}_u$, $\mathbf{B}_v$ and $\mathbf{B}_{v_n}$. We obtain $\mathbf{B}_s$ and $\mathbf{Z}_s$ using the same strategy. $\mathcal{L}_{\mathrm{joint}}$ can be optimised using gradient descent.
\begin{algorithm}[h]
    \caption{Join Learning of $\mathcal{L}_{\mathrm{dom}}$ and $\mathcal{L}_{\mathrm{con}}$.}
    \begin{algorithmic}[1] \small
    \REQUIRE $\mathcal{G}_s, \mathcal{G}_t$: the source and target graphs; $\phi_S=\eta_s(\psi_s(\cdot))$: a source model; $\psi_t$: a target representation learner.
    \STATE $\mathrm{loader}_s$, $\mathrm{loader}_t$ = [$\mathbf{B}^{u,1}_s$,...,$\mathbf{B}^{u, n_s}_s$], [$\mathbf{B}^{u, 1}_t$,...,$\mathbf{B}^{u, n_t}_t$] 
    \FOR{$i$ in n\_epochs}
    \STATE randomise the $\mathrm{loader}_s, \mathrm{loader}_t$ at instance level.
    \FOR{\_ in $\mathrm{min}(n_s, n_t)$}
        \STATE $\mathbf{B}^u_s, \mathbf{B}^u_t \leftarrow \mathrm{src\_loader.next()}, \mathrm{tar\_loader.next()}$
        \STATE$ \mathbf{B}_s, \mathbf{B}_t \leftarrow [\mathbf{B}^u_s, \mathbf{B}^v_s, \mathbf{B}^{v_n}_s],[\mathbf{B}^u_t, \mathbf{B}^v_t, \mathbf{B}^{v_n}_t]$ 
        \STATE $\mathbf{Z}_{s}, \mathbf{Z}_{t} \leftarrow \psi_s(\mathcal{G}_s, \mathbf{B}_s), \psi_t(\mathcal{G}_t, \mathbf{B}_t,)$ \vspace{0.5mm} 
        \STATE update $\psi_t$ w.r.t. $\mathcal{L}_{\mathrm{dom}}(\mathbf{Z}_s, \mathbf{Z}_t)$ //domain alignment
        \STATE update $\psi_t$ w.r.t $\mathcal{L}_{\mathrm{con}}(\mathbf{Z}_t)$ \quad //contrastive learning
    \ENDFOR
    \ENDFOR
    \RETURN {$\psi_t$}
    \end{algorithmic}
    \label{alg:joint}
\end{algorithm}

\subsection{Details of the Representation Learner GraphSAGE}
We choose GraphSAGE \cite{sage} as the representation learner due to its good tradeoff between learning capacity and computational efficiency. Its transformation for generating a given target node $v$'s representation $\mathbf{h}^{k}_{v}$ at the $k$-th layer can be represented as:
\begin{equation}
        \mathbf{h}^{k}_{v} = \sigma(\mathbf{W}^k \cdot \mathrm{AGGR}_k(v, \mathcal{N}(v))),
\end{equation}
where $\mathbf{W}^k$ is the learnable weights and $\mathrm{AGGR}_k(v, \mathcal{N}(v)))$ is the aggregation function from $v$'s immediate neighbours $\mathcal{N}(v)$ for message passing. The mean aggregator is used in our implementation:
\begin{equation}
            \mathrm{AGGR}_k(v, \mathcal{N}(v)) = \mathrm{MEAN}(\{\mathbf{h}^{k-1}_{v}\} \cup \{\mathbf{h}^{k-1}_u, \forall u\in\mathcal{N}(v)\},
\end{equation}
where $\mathbf{h}^{k-1}_{v}$ and $\mathbf{h}^{k-1}_{u}$ are the latent representations of $v$ and $v$'s immediate neighbours' from the previous layer,respectively. For computational efficiency, only some predefined numbers of samples $N_k, \forall k \in K$ are required for the aggregation at each layer.

\end{document}